# Integrated Age Estimation Mechanism

Fan Li[1], Yongming Li[1*, 2], Pin Wang[1], Jie Xiao[1], Fang Yan[1], Xinke Li[1]

(1. School of Microelectronics and Communication Engineering, Chongqing University, Chongqing, 400044, China;

2. Collaborative Innovation Center for Brain Science, Chongqing University, Chongqing 400044, China)

**Abstract**

Machine-learning-based age estimation has received lots of attention. Traditional age estimation mechanism focuses estimation age error, but ignores that there is a deviation between the estimated age and real age due to disease. Pathological age estimation mechanism the author proposed before introduces age deviation to solve the above problem and improves classification capability of the estimated age significantly. However, it does not consider the age estimation error of the normal control (NC) group and results in a larger error between the estimated age and real age of NC group. Therefore, an integrated age estimation mechanism based on Decision-Level fusion of error and deviation orientation model is proposed to solve the problem. Firstly, the traditional age estimation and pathological age estimation mechanisms are weighted together. Secondly, their optimal weights are obtained by minimizing mean absolute error (MAE) between the estimated age and real age of normal people. In the experimental section, several representative age-related datasets are used for verification of the proposed method. The results show that the proposed age estimation mechanism achieves a good tradeoff effect of age estimation. It not only improves the classification ability of the estimated age, but also reduces the age estimation error of the NC group. In general, the proposed age estimation mechanism is effective. Additionally, the mechanism is a framework mechanism that can be used to construct different specific age estimation algorithms, contributing to relevant research.

# 1. Introduction

Age is a high-quality characteristic that profoundly depicts the health states and course of disease. As people age, they progressively lose physical integrity, leading to impaired function and increased susceptibility to diseases such as cardiovascular disease (CVD), diabetes, and Alzheimer's disease (AD) [1]. So age is considered a potentially useful diagnostic marker for disease [2].

The relationship between age and disease process had been reported in some papers [3-8]. In 2013, Gaser et al found early AD patients showed signs of accelerated brain ageing (+10 years) [3]. In the same year, Franke et al found noninsulin-dependent diabetes mellitus subjects have an older brain compared with their chronological age (+4.6 years) [4]. In 2016, Koopman et al assessed the age trend of the hospitalization rate of coronary heart disease (CHD) in the Netherlands from 1998 to 2007, the results showed that the hospitalization rate of CHD decreased in the younger age (< 75 years) , but increased in very old age (>85 years), and the hospitalization age was pushed towards the older age[5]. In 2017, Padur et al observed a majority of patients with systemic arterial hypertension may develop cardiovascular disease and, as age increased, the tendency to develop hypertension also increased [6]. In 2018, Hopper et al pointed out that the absolute breast cancer risk gradient increases with age post menopause, and with underlying familial risk [7].In 2019, Ning et al pointed out factors that are routinely assessed during routine health examinations (gender, age, hemoglobin levels, triglycerides levels, HDL-C, Lp(a) levels, and carotid atherosclerosis) can help identify individuals at higher risk of having Chronic Kidney Disease(CKD) stage 2 [8]. So age estimation is very important.

Age estimation has gained increasing attention worldwide. So far, a number of relevant studies have been published including face age estimation [9-10], dental age estimation[11], skeletal age estimation[12], gestational age estimation [13], wound age estimation[14], death age estimation[15]and brain age estimation[16].Although most of studies show that age quantitative estimation is feasible and effective, and machine learning is an effective tool to mine age information, they do not involve how to use age quantitative estimation for disease detection and diagnosis. Therefore, related researches have begun to focus on this problem. In 2010, Franke et al proposed a framework method of brain age estimation for Alzheimer's disease [17]. In the same year, Dias et al discussed how to estimate the age of teeth for classification of periodontal diseases [18]. In 2015, Moyse et al discussed how to estimate the age of the face for the classification of Alzheimer's disease [19]. In 2016, Liem et al discussed how to use multimodal data to improve brain-based age prediction for Alzheimer's disease[20]. In 2020, Iman et al proposed a brain age estimation framework for Parkinson's disease [21]. The results above show that age estimation has introduced an effective biomarker for detecting and monitoring diseases.

Machine learning has been proved to be an effective way of age estimation such as regression model, feature learning method, instance learning method and age estimation mechanism. The frequently used regression models include support vector regression (SVR) [22], correlation vector regression (RVR) [17], least-squares regression[23-24] , for example, Bukar et al adopted parse partial least squares regression for facial age estimation[10]. Various feature learning methods such as Principal Component Analysis (PCA)[25-26] and manifold learning[27] are applied to establish age estimation models. As to instance learning, Multiple Instance Learning (MIL)[28] and deep instance learning [29] have been employed. Besides, transfer learning [30] and deep learning[31-32] have also been applied to age estimation.

As to age estimation mechanism, all the current methods adopt an estimation mechanism that is to estimate the age by minimizing the difference between the estimated age and real age. For example, In 2016, Lin et al predicted healthy older adult's brain age using artificial neural networks model and the model had mean absolute error (MAE) of 4.29 years[16] .In the same year, Liem et al found that multimodal data improves brain-based age prediction, resulting in a mean absolute prediction error of 4.29 years based on SVR [20]. In 2020, Kuo et al applied an exploratory statistical linear regression model to estimate brain age in large middle-to-late adulthood populations [33]. It is worth noting that the above studies are based on the same mechanism for estimating brain age. The mechanism is to estimate the brain age by minimizing the difference between the estimated age and real age. Firstly, a regression model is used to estimate the age, where the input is the feature data matrix and output is the estimated age. Second, an error function is designed to train the model, and the error function, such as mean absolute error (MAE), is the difference between the estimated age and real age. Third, the optimal estimated age is found by minimizing the error. Therefore, we call this existing age estimation mechanism error orientation model.

There are some problems with this kind of mechanism. Firstly, estimated age varies with disease states,

and the deviation between estimated age and real age varies with disease states. Different disease states correspond to different age deviation. Therefore, it is not suitable to use the real age as the training label. Then, the traditional methods aimed to estimate an age close to the real age by minimizing the error function. Since the real age is not suitable for the training label, the minimization is meaningless for classification of disease or detection of healthy status. Li Y et al introduced age deviation to represent the difference between pathological age and real age to realize estimating pathological age[34-35] . This method builds a regression model based on all classes of samples. The training label is not the real age but the real age plus age deviation. By maximizing the classification accuracy, the optimal age deviation is obtained, and the regression model is trained to converge, so that the optimal estimated age can be obtained. The classification capability of the estimated age has been significantly improved. Therefore, we call this age estimation mechanism deviation orientation model.

However, the study of pathological age estimation method did not consider the age estimation error of normal control (NC) group. The introduction of age deviation into NC group resulted in a larger error between the estimated age and real age of NC group. It is challenging to improve the classification capability of estimated age and reduce the error of age estimation in NC group simultaneously. In order to solve the problem, a novel age estimation mechanism (integrated age estimation mechanism based on Decision-Level Fusion) was proposed here. Firstly, the traditional age estimation and pathological age estimation mechanisms are weighted together. Secondly, their optimal weights are obtained by minimizing the MAE between the estimated age and real age of normal people. With the optimal weight, the integrated age estimation mechanism (IAE) is built. Then, this age estimation mechanism is applied to several public datasets. The results show that the proposed age estimation mechanism achieves a good tradeoff effect of age estimation. It not only improves the classification ability of estimated age, but also reduces the age estimation error of NC group. Since the IAE mechanism is based on deviation and error of age estimation, we call this mechanism integrated (Error and Deviation) orientation model.

This paper is organized as follows: In Section 2, the existing age estimation mechanisms are described. The proposed age estimation mechanism is presented in Section3, followed by the experiment results in Section 4. Section 5 discusses the major innovations and contributions of this paper. Section 6 summarizes this paper.

## 2. Existing Age Estimation Mechanism - TAE

Traditional (existing) age estimation idea and estimated age are called Trad_AgeEstima (TAE) and traditional age respectively. The main idea of TAE mechanism is to estimate the brain age by minimizing the difference between the estimate age and real age. The real age is used as the training label, and the regression model is trained for age estimation by minimizing the MAE between estimated age and real age. Suppose $X=[x_1, x_2, ..., x_N] \in \Re^{d \times N}$ represents the feature set, $Y = [y_1, y_2, ..., y_N]^T$ represents the label matrix of $X$, $\Gamma_{TAE}[\bullet]$ represents the regression model. So the object function $F_1$ of TAE can be expressed as equation (1).

$$F_1 = \min(\frac{1}{N} \sum_{i=1}^{N} |\Gamma_{TAE}[x_i] - y_i|) \qquad (1)$$

The pseudo code of TAE is described as follows (suppose SVR is used for regression)

**Input:** training set $T_r$, verification set $V$, test set $T_e$
**Output:** the traditional age of the test samples.

| | |
|---|---|
| 1 | Initialize: number of test set $N$ |
| 2 | $T_r$ and $V$ are merged, and NC samples are taken as the new training set $T_r^1$; |
| 3 | $T_r^1$ will be sent to the regression model for training, the real age is used as label, and then the trained regression model is obtained; |
| 4 | For $i = 1:N$ |
| 5 | $\quad T_e(i) \to \text{SVR}$ |
| 6 | End |
| 7 | Obtain the traditional estimated age of test samples. |

# 3. The proposed age estimation mechanism - IAE

## 3.1 The Pathological Age Estimation Mechanism-PAE

As the description in section 1, the major problem of TAE is that the error orientation is not helpful for improving the classification capability of the estimated age. The pathological age estimation mechanism proposed by the authors [51-52] and estimated age are called Path_AgeEstima (PAE) and pathological age respectively. PAE mechanism is based on optimal age deviation search orientation. The method uses evaluation criteria of classification capability (such as separability distance and correlation coefficient) to guide the search of age deviation, and searches for the optimal age deviation by maximizing evaluation criteria. Since the deviation between pathological age and real age varies with the state of disease, the deviation is a variable.

Taken two-class classification of disease as example, for class 1, the deviation is set to $p$, which ranges from $p_{\min}$ to $p_{\max}$; the deviation for class 2 is set to $q$, which ranges from $q_{\min}$ to $q_{\max}$. Assuming that the real age of the $i$1th sample in class 1 is $y_{i1}$, the $i$2th sample in class 2 is $y_{i2}$, and the training label of the regression model is not $y_{i1}$, $y_{i2}$ but $y_{i1} + p$, $y_{i2} + q$, respectively.

Firstly, the samples are divided into a training set, validation set and test set. Secondly, the regression model $\Gamma_{PAE}[\bullet]$ is trained using the training set based on the current combination of deviations: $(p,q)$. Then, the input validation set is inserted into the regression model to obtain the estimated ages to calculate the fitness value based on the fitness function (built from evaluation criteria).

The deviations $p, q$ are within $[p_{\min}, p_{\max}]$, $[q_{\min}, q_{\max}]$, respectively, and all the candidate deviations belong to the set of $F_2\{\}|_{p,q}$, The $F_2\{\}|_{p,q}$ is defined as follows: $\{F_2 \in A_{F_2} | A_{F_2} : F_2|_{p,q}, p = p_{\min} : p_{\max}, q = q_{\min} : q_{\max}\}$. In the set, the maximum fitness value $F_{2\max}$ is obtained, and the corresponding optimal deviations are $p_{ma}, q_{ma}$, They are calculated by the following formula (2):

$$[p_{ma}, q_{ma}] = \arg\{F_{2\max}(p_{ma}, q_{ma})\} \qquad (2)$$

The pseudo-code of this mechanism is shown as follows. SVR is used for regression.

---

**Input:** training set $G$, verification set $H$, test set $T$
**Output:** The optimal combination of deviations $(p_{ma}, q_{ma})$ and the pathological age of the test samples.

---

1. Initialization: Current combination of deviations $(p, q)$;
2. According to the current combination of deviations $(p, q)$, modify the age labels of the training sample $G$ and obtain the new training pairs $G^1$. The modification is conducted as follows:
3. For $p \in [p_{\min}, p_{\max}], q \in [q_{\min}, q_{\max}]$
4.     $G^1 \to \{(x_1, y_1 + p), (x_2, y_2 + q)\}$;
5.     $G^1$ will be sent to the SVR for training, and the SVR model is obtained after training;
6.     $H \to SVR$, obtain the estimated ages, and calculate and store the fitness value of $F_2|_{p,q}$;
7.     If $p = p_{\max}, q = q_{\max}$
8.         Select the maximum value $F_{2\max}$ from the stored $F_2|_{p,q}$ and its corresponding combination of deviations $(p_{ma}, q_{ma})$ and the optimal model $SVR_{optimal}$;
9.         Quit the circle.
10.     End
11. End
12. $T \to SVR_{optimal}$, obtain the pathological age of test samples.

---

## 3.2 Integrated_AgeEstima Mechanism Combining Path_AgeEstima and Trad_AgeEstima

Integrated age estimation mechanism and estimated age are called Integrated_AgeEstima (IAE) and integrated age respectively. IAE mechanism is designed based on the optimal deviation search method of kernel regression. The TAE and PAE mechanisms are weighted fusion by using the weighted integration optimization mechanism(see Figure1).

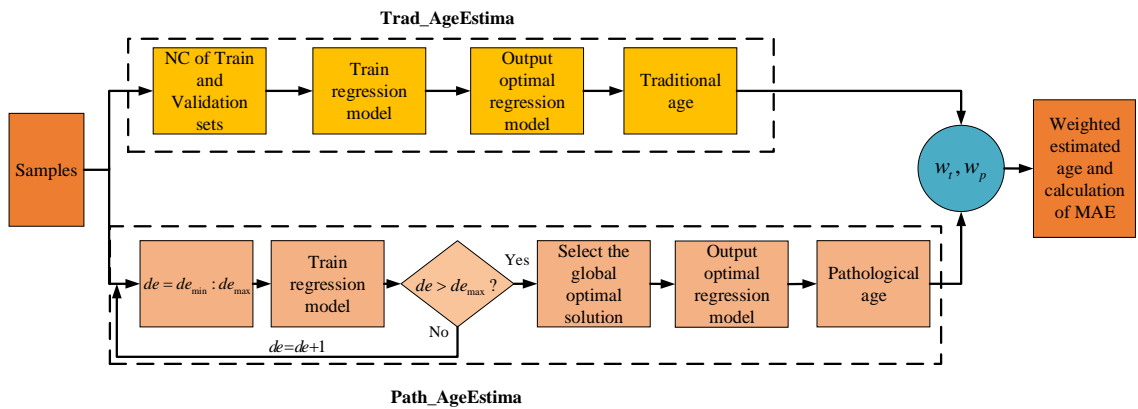

Fig1.The structure of weighted integration module

The range of traditional age weight $w_t$ is $[w_{t\min}, w_{t\max}]$, and pathological age weight $w_p$ is

$[w_{p\min}, w_{p\max}]$. Therefore, the objective function of weighted integration mechanism optimization can be expressed as formula (3)

$$\min_{w_t, w_p} \frac{1}{N_{NC}} \sum_{i=1}^{N_{NC}} \left| \left( w_t * \Gamma_{\mathbf{TAE}}[x_i] + w_p * \Gamma_{\mathbf{PAE}}[x_i] \right) - y_i \right|$$
$$\text{s.t.} \quad w_t + w_p = 1 \qquad (3)$$
$$w_t \in [w_{t\min}, w_{t\max}], w_p \in [w_{p\min}, w_{p\max}]$$

Taking $\hat{y}_{ti} = \Gamma_{\mathbf{TAE}}[x_i], \hat{y}_{pi} = \Gamma_{\mathbf{PAE}}[x_i]$, the modified objective function is given by:

$$\min_{w_t, w_p} \frac{1}{N_{NC}} \sum_{i=1}^{N_{NC}} \left| \left( w_t * \hat{y}_{ti} + w_p * \hat{y}_{pi} \right) - y_i \right| \qquad (4)$$

where $\hat{y}_{ti}$ represents the traditional age of $i$th sample in the NC group, $\hat{y}_{pi}$ represents the pathological age of $i$th sample in the NC group. By optimizing (4), the optimal weight $w_{t\_opt}, w_{p\_opt}$ are obtained, and the final estimated age is shown formula (5)

$$\hat{y}_i = w_{t\_opt} * \hat{y}_{ti} + w_{p\_opt} * \hat{y}_{pi} \qquad (5)$$

The pseudo-code of this proposed mechanism (IAE) is shown as follows (SVR is used for regression).

---

**Input:** training set $T_r$, verification set $V$, test set $T_e$
**Output:** The optimal combination of weights $(w_{t\_opt}, w_{p\_opt})$ and the integrated age of the test samples

1. Initialization: $de = de_{\min}$, $w_t = w_{t\min}$
2. Merge $T_r$ and $V$, divide them into new training set $T_r^1$, verification set $V^1$ and test set $T_e^1$;
3. Get the traditional age_1 of $T_e^1$ with TAE and store the optimal model $\text{SVR}_{opt}^t$;
4. Get the pathological age_1 of $T_e^1$ with PAE and store the optimal model $\text{SVR}_{opt}^p$;
5. For $w_t \in [w_{t\min}, w_{t\max}]$, $w_p = 1 - w_t$
6.     Obtain weighted estimated age by $w_t$ * traditional age_1 + $w_p$ * pathological age_1;
7.     Calculate the MAE between the weighted age and the real age of the NC group in $T_e^1$, and preserve the MAE value of $\mathbf{M}|_{w_t, w_p}$;
8.     If $w_t = w_{t\max}, w_p = w_{p\min}$
9.         select the minimum value $\mathbf{M}_{\min}$ from the stored $\mathbf{M}|_{w_t, w_p}$ and its corresponding weights $(w_{t\_opt}, w_{p\_opt})$;
10.     End
11. End
12. $T_e \to \text{SVR}_{opt}^t$, obtain the traditional age of test samples;
13. $T_e \to \text{SVR}_{opt}^p$, obtain the pathological age of test samples;
14. Weight the traditional ages and pathological age by the optimal weight $(w_{t\_opt}, w_{p\_opt})$ to get the integrated age of the test set samples.

# 4. Results

## 4.1. Experimental condition

To demonstrate the effectiveness of the proposed mechanism, some experiments are conducted including parameters analysis, reliability analysis and comparison of different age estimation mechanisms. Parameter analysis is mainly to consider the optimal kernel function, feature reduction method and weights. Reliability analysis is to verify the reliability of the proposed mechanism. Comparison experiments are organized to show the proposed mechanism can achieve a good tradeoff effect on age estimation.

Seven public datasets available from UCI, ADNI and Kaggle dataset repository are chosen for verification, they are Heart disease (HD), Diabetes mellitus (DM), Alzheimer's disease (AD), Breast Cancer Coimbra (BCC), Indian Liver Patient Dataset (ILPD), Coronary artery disease (CAD), and Chronic kidney disease (CKD).

1) HD dataset: The total number of samples was 297, consisting of two classes of samples: NC and DG (Disease Group), the number of NC samples was 160, the number of DG samples was 137. And it contains 13 features, including age, type of chest pain, resting blood pressure, serum cholesterol and category labels.
2) DM dataset: The total number of samples was 392, which consists of 160 NC samples and 132 DG samples, and it includes 9 features, such as age, blood sugar, blood pressure, sebum thickness and category label.
3) AD dataset: The total number of samples was 287, which consists of 162 NC samples and 125 DG samples, and it includes 4 features: age, left and right volume of hippocampus and category label.
4) BCC dataset: The total number of samples was 116, consisting of two classes of samples: NC and DG (Disease Group), the number of NC samples was 52, the number of DG samples was 64. and it contains 10 features, such as age, BMI, Glucose and so on.
5) ILPD dataset: The total number of samples was 583, consisting of two classes of samples: NC and DG (Disease Group), the number of NC samples was 167, the number of DG samples was 416. and it contains 9 features.
6) CAD dataset: The total number of samples was 303, consisting of two classes of samples: NC and DG (Disease Group), the number of NC samples was 87, the number of DG samples was 216. and it contains 21features.
7) CKD dataset: The total number of samples was 215, consisting of two classes of samples: NC and DG (Disease Group), the number of NC samples was 128, the number of DG samples was 87. and it contains 11 features.

A basic information of each dataset is given in Table 1.

Table 1 Basic information of each dataset

| Dataset | NC/DG number | Attribute | NC/DG Age ranges (years) | NC/DG Mean age (years) | NC/DG Age standard deviation |
|---|---|---|---|---|---|
| HD | 160/137 | 13 | 29-74/35-77 | 52.31/56.76 | 9.09/7.90 |
| DM | 160/132 | 9 | 21-63/21-60 | 28.42/35.94 | 8.99/10.64 |
| AD | 162/125 | 4 | 62-90/57-90 | 76.36/75.69 | 5.23/7.10 |
| BCC  | 52/64  | 9  | 28-89/34-86 | 58.08/56.67 | 18.96/13.49 |
| ILPD | 167/416| 9  | 4-85/14-90  | 41.24/46.15 | 16.99/15.65 |
| CAD  | 87/216 | 21 | 40-63/41-81 | 53.06/61.25 | 9.32/9.88   |
| CKD  | 128/87 | 11 | 12-80/6-90  | 47.39/58.77 | 15.21/13.4  |

In this paper, the experimental operating system platform is the Windows10, 64-bit operating system, and the memory size is 8 GB. The data processing is completed in MATLAB, version 2018b. For the sake of fairness, other relevant parameters of SVR except kernel function are set to default values: the saliency level $\alpha$ of Pvalue is set to 0.05; the characteristic weight threshold of Relief is set to 500; and the precision of PCA is set to 0.9999. The other related parameters involved in the algorithm are all default values. The range of $p$ and $q$ is set to [-10,10], and the range of $w_t$, $w_p$ is set to [0,1]. By hold-out cross-validation method, the samples are randomly divided into training set, validation set and test set 30 times, yielding 30 groups of samples, and all the following experimental results are the statistical average results.

## 4.2. Comparison with existing age estimation mechanism

In order to further verify the effectiveness of the proposed mechanism, TAE and PAE mechanism with $\lambda_1$, $\lambda_2$ are compared here. The estimated age of test samples is obtained with IAE, TAE, PAE with $\lambda_1$, $\lambda_2$, and then fitness values and MAE are calculated.

For all datasets, fitness and MAE values obtained by above mechanisms with NoFC are shown in table 2.

Table2. Comparison of different mechanisms for seven datasets

| Mechanisms | IAE | | TAE | | PAE with $\lambda_1$ | | PAE with $\lambda_2$ | |
|---|---|---|---|---|---|---|---|---|
| Datasets | $\lambda_1$/MAE | $\lambda_2$/MAE | $\lambda_1$/MAE | $\lambda_2$/MAE | $\lambda_1$/MAE | $\lambda_2$/MAE | $\lambda_1$/MAE | $\lambda_2$/MAE |
| HD   | **0.735/7.832**  | **0.629/7.832**  | 0.529/8.587 | 0.533/8.038 | 0.682/8.587  | - | - | 0.620/8.587  |
| DM   | **0.310/4.961**  | **0.473/4.961**  | 0.287/5.422 | 0.459/5.422 | 0.303/5.921  | - | - | 0.467/5.921  |
| AD   | **0.573/3.898**  | **0.581/3.898**  | 0.541/3.900 | 0.543/3.900 | 0.654/4.352  | - | - | 0.619/4.352  |
| BCC  | **0.128/16.077** | **0.159/16.437** | 0.039/16.682| 0.135/16.682| 0.128/16.542 | - | - | 0.155/16.912 |
| ILPD | **0.055/15.199** | **0.197/15.199** | 0.034/15.201| 0.150/15.240| 0.054/16.172 | - | - | 0.195/16.172 |
| CAD  | **0.089/7.269**  | **0.239/7.232**  | 0.026/7.281 | 0.128/7.281 | 0.087/9.055  | - | - | 0.239/10.207 |
| CKD  | **6.101/11.421** | **0.920/11.421** | 0.808/11.819| 0.451/11.823| 6.156/14.209 | - | - | 0.921/14.021 |

As seen from table 2, for all datasets, IAE had better separability and MAE (see the boldface type) than TAE and PAE respectively in most cases. In order to further show the advantages of the proposed mechanism intuitively, figure 4 shows the bar graph results measured by different age estimation mechanisms in table 2 taking HD and DM as examples.

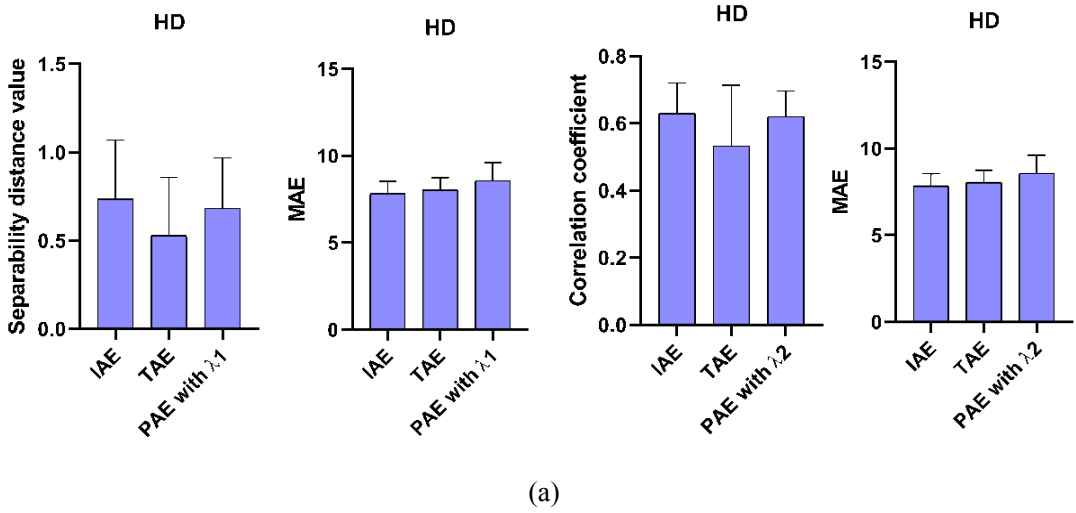

(a)

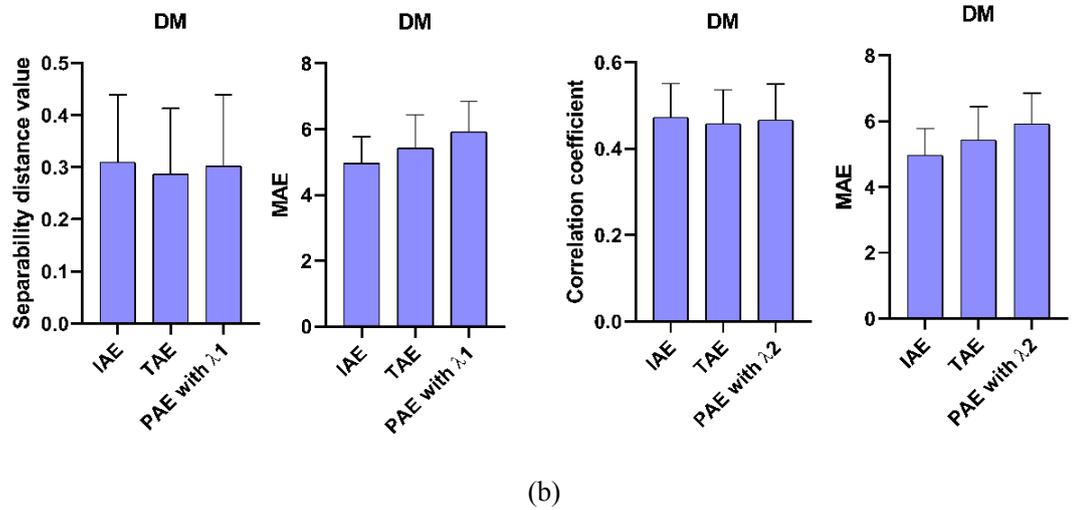

(b)

Fig 2. Results of the age estimation mechanisms on different datasets: (a) is the comparison of separability distance, correlation coefficient and MAE values for HD; (b) is the comparison of separability distance, correlation coefficient and MAE values for DM;

Compared with TAE and PAE, IAE improves the classification ability of estimated age and reduces the age estimation error of NC group for all datasets. Although the classification ability of estimated age in AD and CKD decreased to some extent, the MAE is significantly reduced. In general, IAE can effectively consider the classification ability of the existing age estimation mechanisms and the age estimation error of NC group at the same time, so it can achieve an excellent tradeoff effect.

## 4.3. Visualized analysis of separability of estimated age on test samples

For further showing the advantages of IAE. Figure3 shows the distribution of the test samples with age-features measured by different methods in table2. Here, the real age and estimated age are used for the test samples' features as two dimensions, the x-axis is real age, and the y-axis is estimated age.

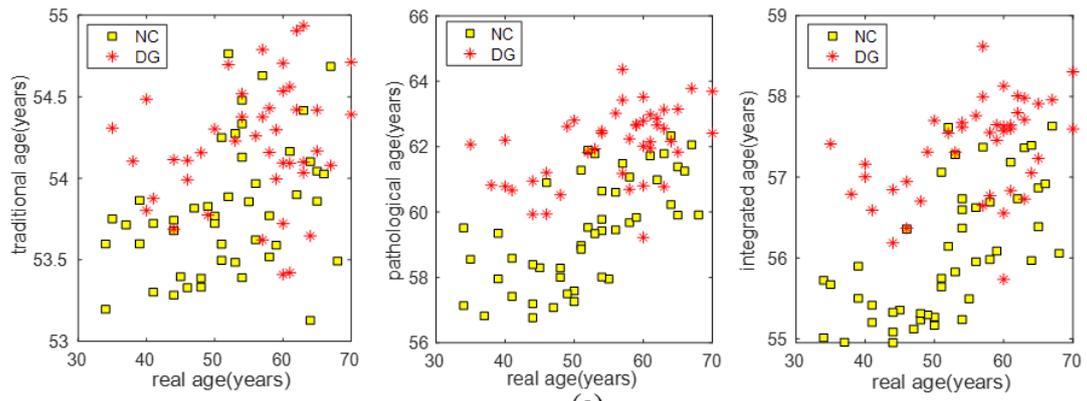

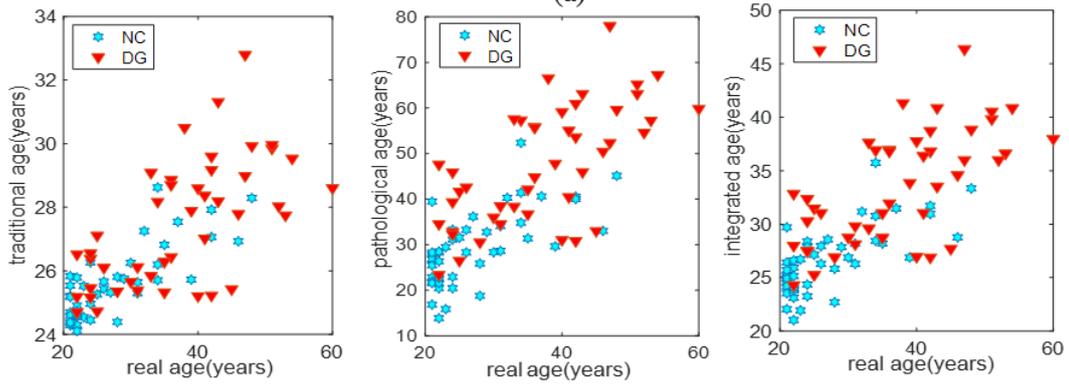

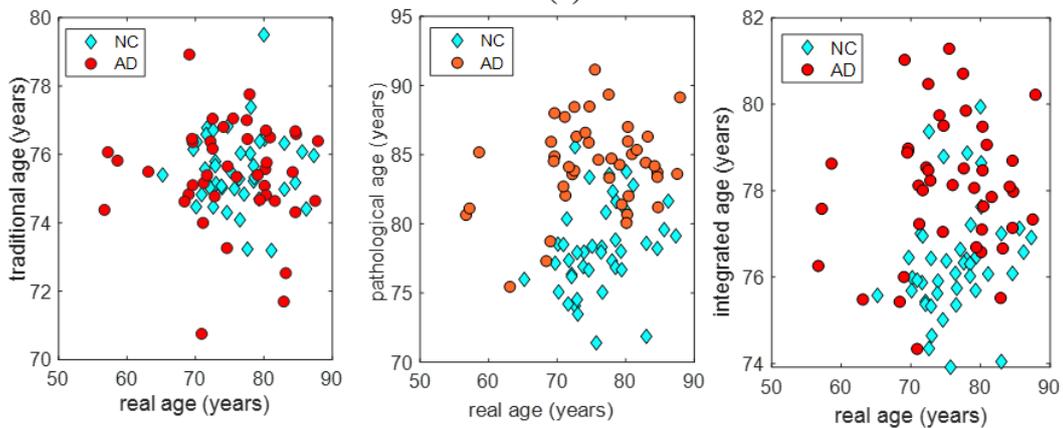

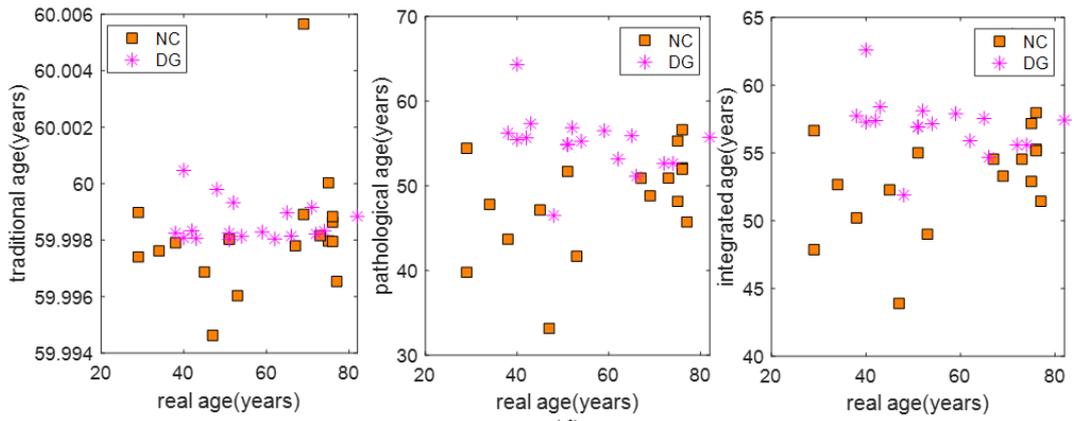

(d)

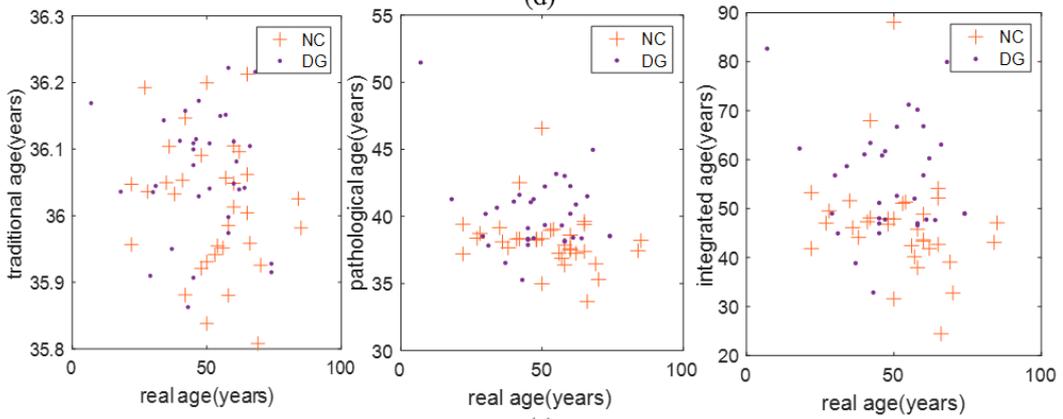

(e)

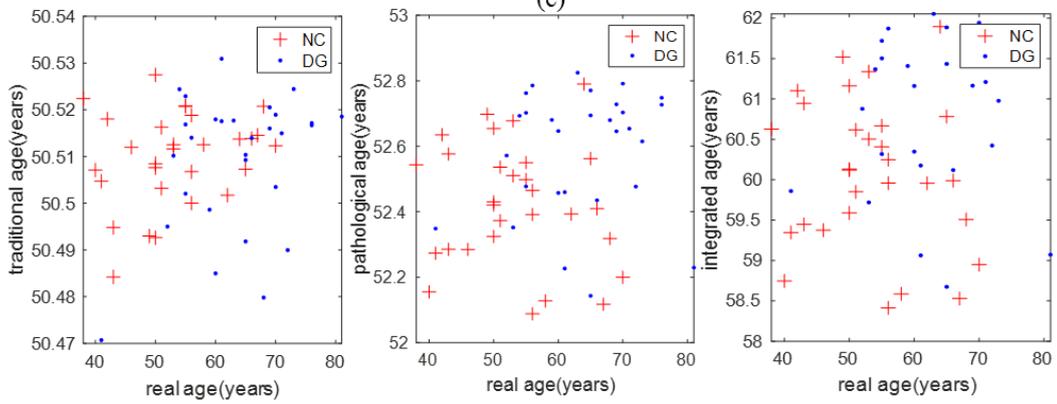

(f)

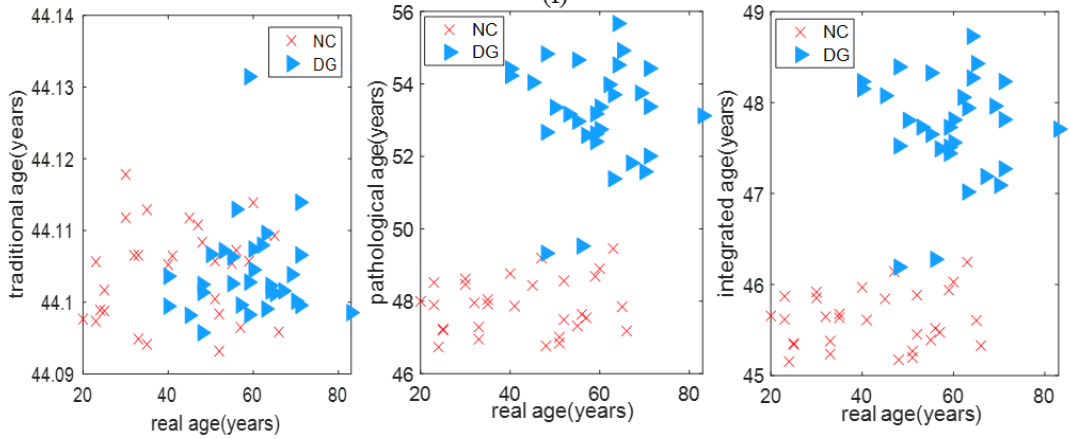

(g)

Fig3. The visualized comparison of separability based on $\lambda_1$ with TAE,PAE and IAE respectively: (a) is the comparison of three mechanisms on HD for NC-DG;(b) is the comparison of three mechanisms on DM for NC-DG ;(c) is the comparison of three mechanisms on AD for NC-AD;(d)is the comparison of three mechanisms on BCC for NC-DG;(e) is the comparison of three mechanisms on ILPD for NC-DG;(f) is the comparison of three mechanisms on CAD for NC-DG;(g) is the comparison of three mechanisms on CKD for NC-DG

Figure 3 denotes the distribution of test samples with TAE, PAE as well as IAE for seven datasets. As shown in figure 3(a), for HD, it can be seen that the test samples with estimated age become separable using TAE. However, it is not easy to distinguish the two classes of samples. The test samples with estimated age become more separable using PAE. Finally, when IAE is used, the test samples with estimated age are most separable, IAE becomes easiest to distinguish the two classes of samples compared with the former two mechanisms. Besides, for the IAE, the MAE is the smallest. The similar case is true for other six datasets. Thus, the proposed IAE mechanism has the best classification capability, it is superior to other two mechanisms (TAE, PAE).

## 4.4. Reliability analysis

Intraclass correlation coefficient (ICC) is used for evaluating the reliability of TAE and IAE. The results are shown in Table 3 and Figure 4. Figure 4 denotes the distribution of NC samples with TAE and IAE for all datasets.

Table3 Reliability estimation of TAE and IAE

| Mechanisms<br>Dataset | TAE(ICC) | IAE(ICC) |
|---|---|---|
| HD | 0.509 | **0.600** |
| DM | 0.782 | **0.808** |
| AD | 0.720 | **0.772** |
| BCC | 0.451 | **0.507** |
| ILPD | 0.352 | **0.430** |
| CAD | 0.605 | **0.623** |
| CKD | 0.292 | **0.300** |

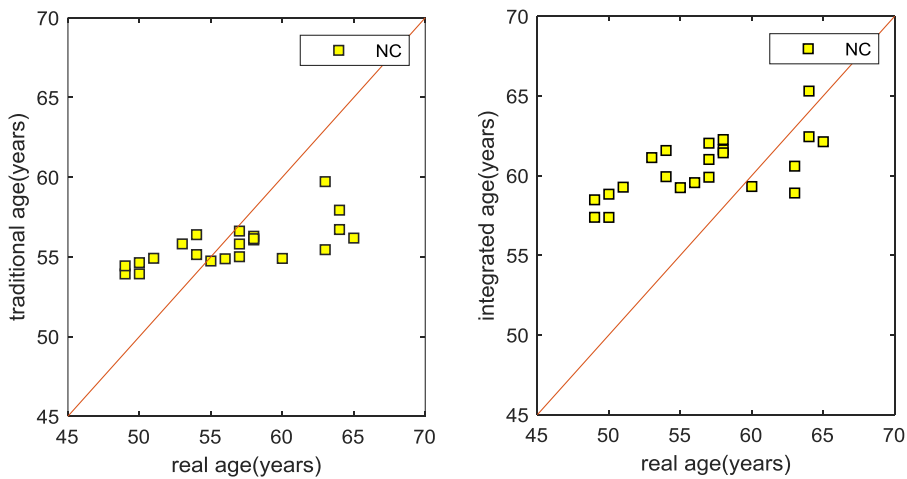

(a)

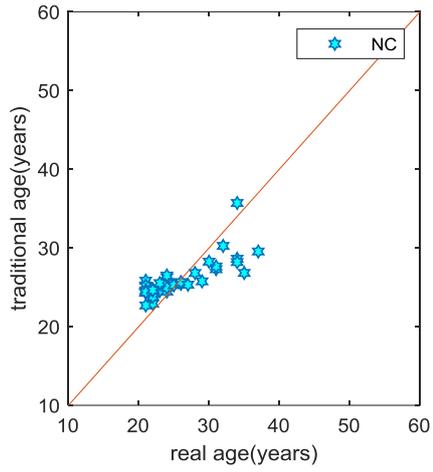
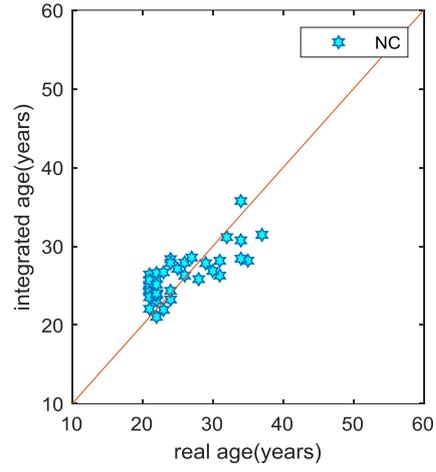

(b)

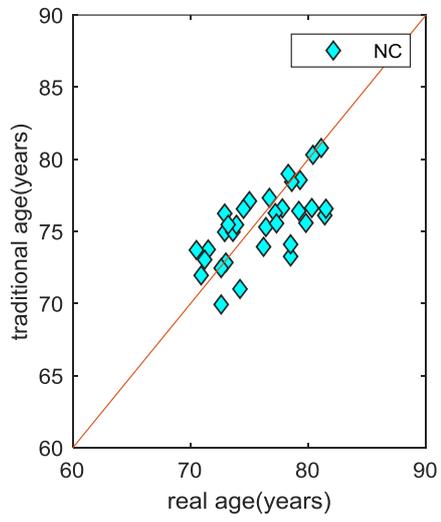
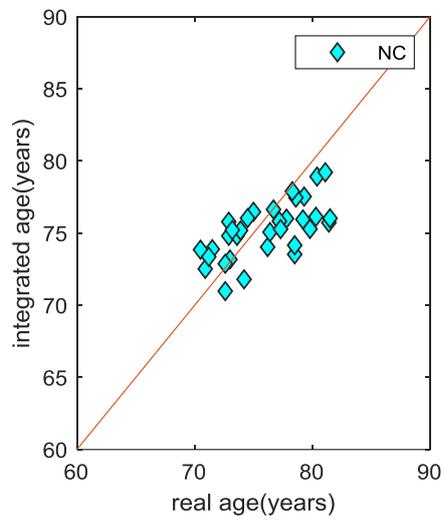

(c)

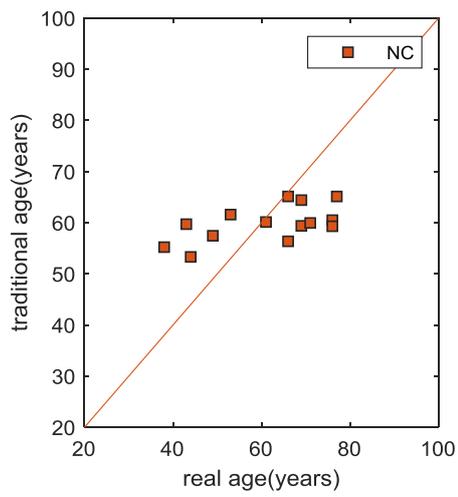
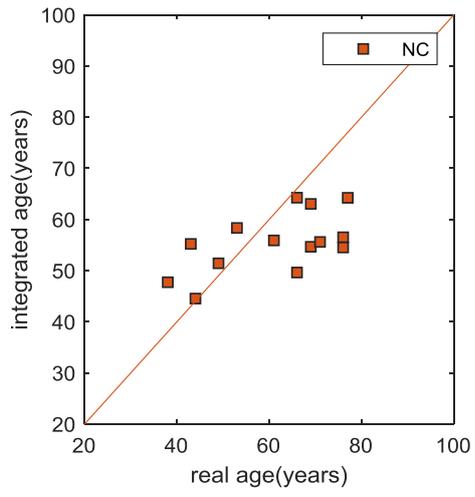

(d)

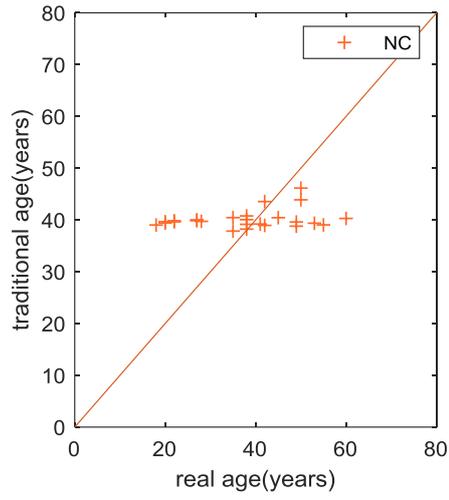
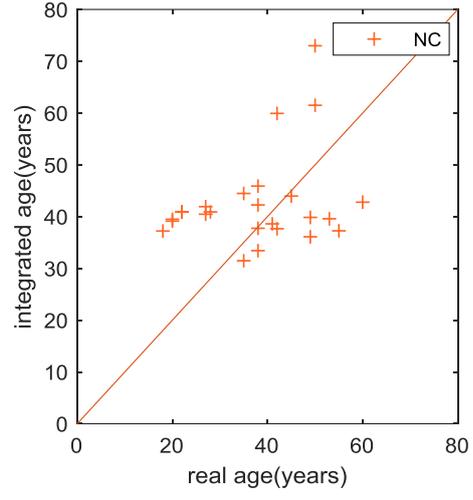

(e)

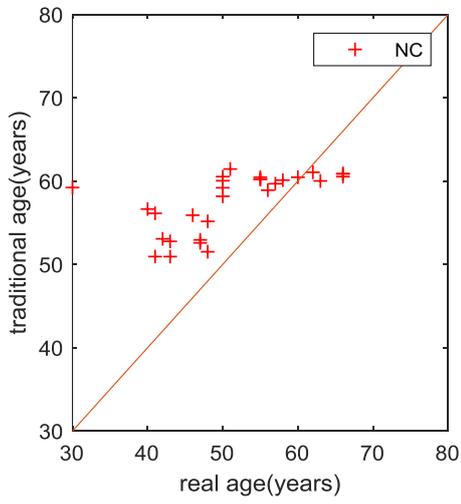
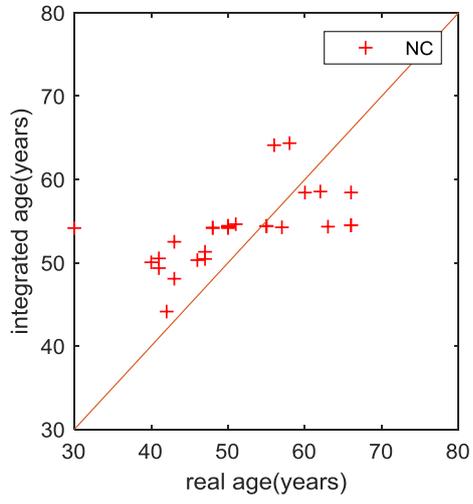

(f)

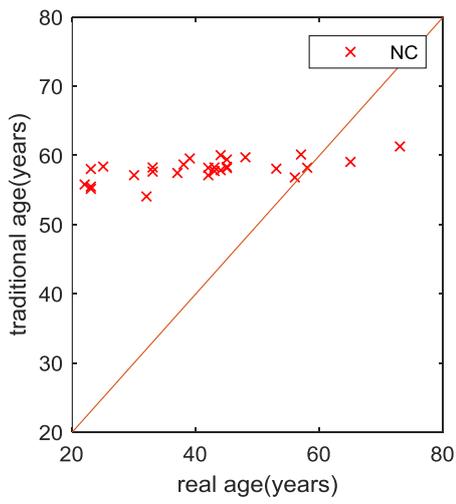
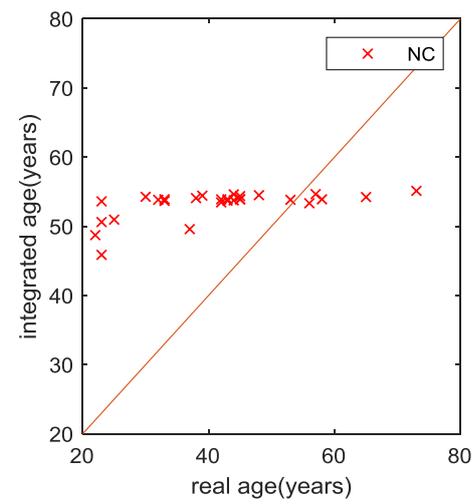

(g)

Fig 4 Reliability of TAE and IAE for datasets, the left column represents TAE, and the right column represents IAE: (a) is for HD;(b) is for DM; (c) is for AD;(d) is for BCC;(e) is for ILPD;(f) is for CAD;(g) is for CKD

Table 3 records the ICC of TAE and IAE. As can be seen, the bold fonts denote ICC values with IAE are more excellent than TAE. For DM, ICC values are 0.808, 0.782 and for AD, those are 0.772, 0.720 for IAE and TAE respectively. Figure 4 shows the relationship between the estimated age and real age for two mechanisms. In figure 4, it can be found that IAE outperforms TAE, in terms of correlations between integrated age and real age. So, in this case, IAE presents better performance than TAE.

## 5. Discussion

Age plays an important role in the diagnosis of diseases. The traditional age estimation mechanism can distinguish the different states of diseases, and it helps improve classification accuracy. The main idea is to estimate the age by minimizing the distance between the estimated age and real age. This idea is not in accordance with the fact that some diseases processes are forms of accelerated aging. So the authors solved this problem based on pathological age by maximizing the classification accuracy before. They introduced age deviation to represent the difference between pathological age and real age to realize the method of estimating pathological age, but the pathological age estimation mechanism not well considered reducing the estimation error of the normal control. The two age estimation mechanisms can be looked as error orientation model and deviation orientation model respectively.

So, an integrated age estimation mechanism based on decision-level fusion of error orientation and deviation orientation is proposed to solve the problem above. Firstly, samples are divided into three parts: training set, validation set and test set. Secondly, traditional and pathological age are obtained based on training and validation set. Thirdly, the weight and MAE of NC are introduced. Fourthly, based on the weight candidate, the corresponding MAE of NC is obtained. Fifthly, the weight is optimized by minimizing the MAE.

The popular regression machine SVR is used as age estimation model, and separability distance and correlation coefficient are used as evaluation criteria. To further show the advantage of the proposed IAE mechanism, the performance of the TAE and IAE is compared. For the same test set, it can be seen IAE had the highest separability except for AD, CKD datasets and smallest MAE for all datasets. Visualized analysis of separability capability and MAE on test samples is also provided. Based on the comparison for all datasets, IAE has the highest separability and smallest age estimation error than the TAE.

Seven public tabular datasets are used for verification of the proposed mechanism, and they are HD, DM, AD, BCC, ILPD, CAD, and CKD. Although some literature uses image or voice data for age estimation, it requires raw data preprocessing such as feature extraction, so the various preprocessing methods affect the comparison of age estimation. For fair comparison, the tabular public datasets which contain age information are involved directly.

This paper proposed a new idea (or mechanism) for age estimation. The main innovations and contributions of this paper can be described as follows.
(1) The current age estimation mechanisms for the diagnosis of disease are based on the same idea – error orientation model (i.e TAE). This paper proposed a new idea to replace it rather than proposed a new concrete method.
(2) This proposed idea not only improves the ability of age estimation classification, but also reduces the age estimation error of NC group.
(3) A novel age estimation mechanism fusion is designed for the first time by combining the error orientation and deviation orientation, thereby constructing an integrated age estimation mechanism.

(4) The existing age estimation mechanism is a special case of the proposed integrated age estimation mechanism.

Although some achievements have been made in this paper, future work remains: First, in this study, the proposed age estimation mechanism is mainly based on machine learning, so future work will focus on the introduction of deep learning into age estimation. Second, other regression models can be considered for age estimation.

# 6.Conclusion

Many studies support the relationship between age and health and age estimation based on machine learning has received wide attention. TAE mechanism focuses estimation age error, but ignores that there is a deviation between the estimated age and real age due to disease. PAE mechanism introduces age deviation to solve the above problem and improves classification capability of the estimated age significantly. However, it does not consider the age estimation error of NC group and results in a larger error between the estimated age and real age of NC group. In this paper, an integrated age estimation mechanism based on Decision-Level fusion of error and deviation orientation model is proposed. In the experimental section, seven public datasets that include age information are used for verification. The experimental results show that IAE mechanism achieves a good tradeoff effect of age estimation. It not only improves the classification ability of the estimated age, but also reduces the age estimation error of the NC group. The advantage of the proposed mechanism is apparent for different datasets. The proposed mechanism is not restricted to a concrete regression model or feature learning method and can be applied to improve any existing age estimation algorithm. Therefore, this study is heuristic to relevant researchers.

# Acknowledgements

This work was supported in part by the National Natural Science Foundation of China (NSFC) under Grant 61771080, in part by the Fundamental Research Funds for the Central Universities under Grant 2019CDQYTX019, Grant 2019CDCGTX306 and 2019CDQYTX017, in part by the Basic and Advanced Research Project in Chongqing under Grant cstc2018jcyjAX0779, cstc2020jcyj-msxmX0523 and cstc2020jscx-msxm0369, in part by the Research on tumor metastasis and individualized diagnosis and treatment in Cancer Hospital Affiliated to Chongqing University Chongqing Key Laboratory Open Project Fund General Project.# Conflict of interest

The authors declare that they have no conflicts of interest related to this work.